\pdfoutput=1
\documentclass[11pt,a4paper]{article}
\usepackage[hyperref]{acl2021}
\usepackage{times}
\usepackage{latexsym}
\usepackage{graphicx}

\usepackage{microtype}

\aclfinalcopy 

\title{Visualization Techniques to Enhance Automated Event Extraction}

\author{Sophia Henn \\University of Notre Dame, shenn@alumni.nd.edu\\ \bf Abigail Sticha \\University of Notre Dame, asticha@nd.edu\\ \bf Timothy Burley \\University of Notre Dame, tburley@alumni.nd.edu\\ \bf Ernesto Verdeja \\University of Notre Dame, everdeja@nd.edu\\ \bf Paul Brenner \\University of Notre Dame, pbrenne1@nd.edu}

\date{}

\begin{document}
\maketitle
\begin{abstract}
Robust visualization of complex data is critical for the effective use of NLP for event classification, as the volume of data is large and the high-dimensional structure of text makes data challenging to summarize succinctly. In event extraction tasks in particular, visualization can aid in understanding and illustrating the textual relationships from which machine learning tools produce insights. Through our case study which seeks to identify potential triggers of state-led mass killings from news articles using NLP, we demonstrate how visualizations can aid in each stage, from exploratory analysis of raw data, to machine learning training analysis, and finally post-inference validation.
\end{abstract}

\section{Introduction}

As the number, capacity, and complexity of automated event extraction tools and techniques grow, researchers must continue to enhance their capability to quickly visualize massive data sets and incremental artifacts at key points in the event classification and extraction workflow.  In service to this need we present a summary of related visualization tools followed by visualization techniques that our team finds critical to enhancing the automation efficiency and event extraction validation.

\subsection{Related Works}

Visualization plays a unique role in the comprehension of linguistic structures. A large body of literature is dedicated to exploring how text data can be best summarized and represented in lower dimensions. These studies describe how visualizations including network representations, dendrograms, radar charts, and vector maps of word embeddings are suitable to communicate syntactical and semantic relationships within text \cite{santini-etal-2020-visualizing1,DRIEGER20134,handle:10.2312:cgvc20181219}.

Visualization similarly supports a range of machine learning tasks with text data including topic modeling, information retrieval, sentiment analysis and classification \cite{textflow,treemapinfo,SocialHelixVA,ClassFeatureInsight}. Since raw text usually undergoes several preprocessing steps to transform it into a usable input for different models, visualization of the text at different stages helps to provide insight into the model and call attention to features that might negatively impact model performance. \citet{8356097} summarize a range of common techniques suitable for each type of task, including typographic visualizations, spatial projections, and topological maps. These visualizations generally take tokenized text or metadata as inputs and communicate insights specific to the task at hand (summarize the text, identify outliers, or explore differences in training data for supervised learning algorithms).

Visualization is also increasingly being leveraged to aid in model selection and validation \cite{DBLP:journals/corr/LiuWLZ17,Lutrendsanalytics}. \citet{doi:10.1177/1473871620904671} overview visualization techniques used for model interpretation and refinement, and \citet{GARCIA201830} emphasize the use of visualizations to aid in parameter tuning and feature selection of neural networks. \citet{doi:10.1177/1473871620922166} use visual encodings of several dimensionality reduction techniques to compare human and machine estimates of clusters in a dataset. This paper summarizes how visualization techniques have been useful to garner insights into our complex data and particular event classification task.

\subsection{Triggers of Mass Killings Case Study}
Our case study aims to utilize statistical and NLP tools for a systemic analysis of triggers of state-led mass killings (ToMK). Peace and conflict researchers have identified several large-scale structural conditions that make state-led mass killings more likely, such as political instability, a history of violence against vulnerable groups, radical political ideologies, and autocratic or authoritarian governments.
Several existing models aim to forecast the escalation of violent conflict \cite{VIEWS,goldsmithforecast,goldstoneForecast}. However, the timing of mass killing onset is less understood. Analysts have identified several plausible trigger-type events including coups, assassinations, protests/riots, armed conflict escalation, cancelled elections, neighboring conflict spillover and others, but little systematic analysis beyond specific country case studies has been conducted to examine whether, and if so when, these events actually trigger mass killings.

\begin{figure}
  \centering
  \includegraphics[width=0.95\linewidth]{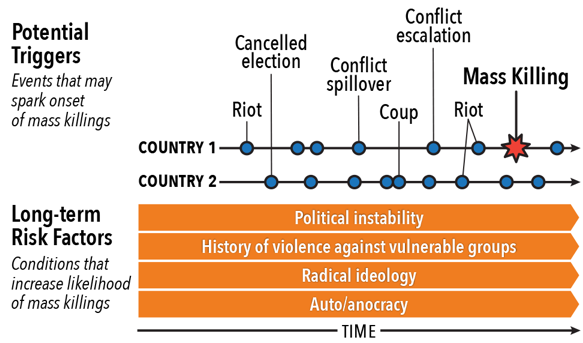}
  \caption{Potential triggers and structural conditions for state-led mass killings.}
  \label{figs:TrigCond}
\end{figure}

This project canvasses a large number of news sources to identify and examine the occurrence of nine trigger-type events across both countries that have had a mass killing event and those that have not between 1989-2017, and analyzes under what conditions - and in what patterns and sequences - certain trigger-type events increase the probability of mass killings (Figure \ref{figs:TrigCond}). It seeks to bring greater specificity and understanding about the timing of state-led mass killings, which is of interest to both scholars and peacebuilders.

The ToMK project utilizes SVMs and BERT \cite{liu2010study,devlin2019bert, geron2019hands}, to identify patterns in the text data and classify news articles as a positive or negative instance of a trigger event. The SVM script preprocesses the raw text, vectorizes each article using a tf-idf weighting scheme, and defines a classification hyperplane to separate the classes of articles.  We then use this kernel to classify unlabeled data. 

\subsection{Data Specification}
This paper focuses on the coup/attempted coup trigger. The coup data consists of the text of news articles retrieved via LexisNexis queries based on several search parameters which include a date filter from 1989-2017, a source filter for our list of 20 sources, and keywords including the word “coup” or related terms.  The articles are compiled into CSV files with columns for the article text, date of publication, unique document id, and country label based on the keyword in the search.  The calculations for the maximal corpus yield a worst-case estimate of 1.8 million articles across all countries.  The visualizations in this paper use a subset of this corpus (``select countries corpus") that contains the 69 countries with identified state-led mass killing events between 1989-2017. This select countries corpus contains 602,939 articles.  A subset of this corpus isolates the two-year window preceding each mass killing event (``dependent space corpus") and contains 51,488 articles.  The machine learning training set consists of 551 articles retrieved in the same manner with an additional CSV column of human-coded labels indicating whether the article is describing a qualified coup event.

\section{Raw Data Visualizations}
\label{sec:Pre-Class}

\begin{figure}
  \centering
  \includegraphics[width=0.95\linewidth]{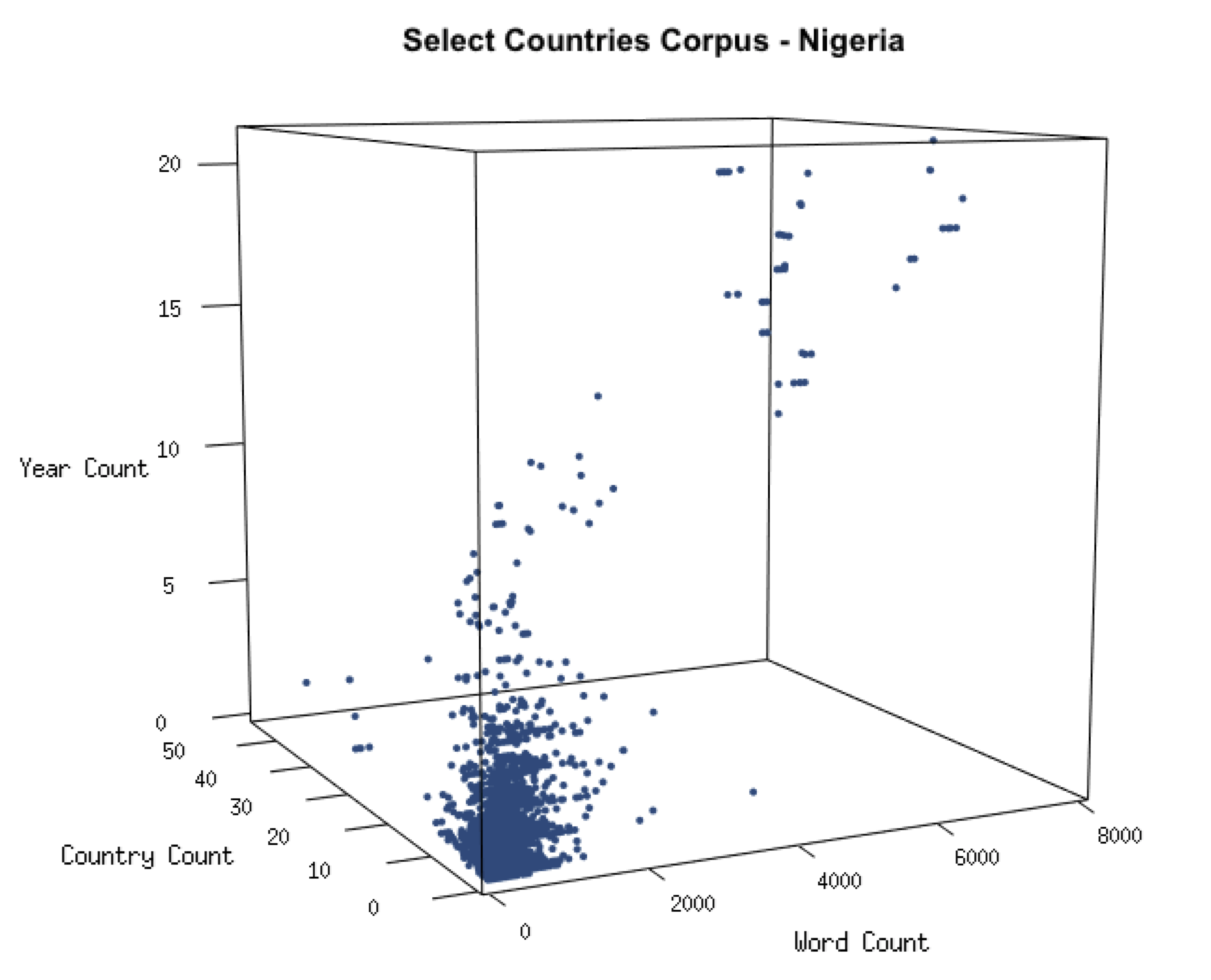}
  \caption{Correlation plot of article characteristics for insight into potentially problematic articles and their frequency across the corpus.}
  \label{figs:RAW}
\end{figure}

Initial, exploratory visualizations are useful to summarize large corpora and confirm that the search parameters were successful in returning events of interest.  Necessary initial preprocessing includes transforming text to lowercase, tokenization, and removing non-alphanumeric characters.

To describe the shape of the corpus, histograms of word counts can be informative by communicating the variation in article length. It is useful to consider what lengths one expects given the genre of text in the corpus. Social media posts, news articles, or novels all carry different expectations of lengths, so histograms can be used to view this distribution and any outliers. In our particular case study we are interested in news articles, so we expect most text to fall under 2000 words. Histograms can also confirm that the data was loaded properly; some programs, like Microsoft Excel, have default maximum character counts per cell, so a cutoff around a particular word count on the histogram could signal an issue with data importing.

Word clouds or variations like scattertext are also popular to summarize the word content of the corpus. However, while word clouds can be descriptive, they can be misleading regarding what words are significant towards event classification tasks. Section 3 will visualize significant word tokens towards classification in our study after applying the tf-idf weighting scheme.



News articles can take on many structures, with some more complicated than a description of a single event that occurred recently.  The presence of articles with complex formats impacts our event classification task in two primary ways. Firstly, if an article describes a single coup event along with other world events, there will be some minimal language describing a coup but a predominance of unrelated words; this may impact the weights of keywords and lead to more frequent misclassification. Secondly, the country or date labels for the coup event may be incorrect since the labels are extracted from the keyword search parameters for that query. News digests are an example of these types of articles that summarize many events around the world for a given week or month, and they are published among standard news articles by several of our sources.

We identified three characteristics associated with news digests: higher word counts, counts of unique country names, and counts of unique years.  For each article, the word count, country count and year count was calculated and all articles were plotted for a particular country.  Figure 2 plots these characteristics for articles with the country label Nigeria.  This visualization confirms the correlation between these characteristics, and display how common news digests may be within our corpus.  Articles above a percentile for the three characteristics can be marked for human reader validation after classification or thrown out of the corpus before feature extraction and classification.


\section{Machine Learning Visualizations}

 \begin{figure}
  \centering
  \includegraphics[width=0.95\linewidth]{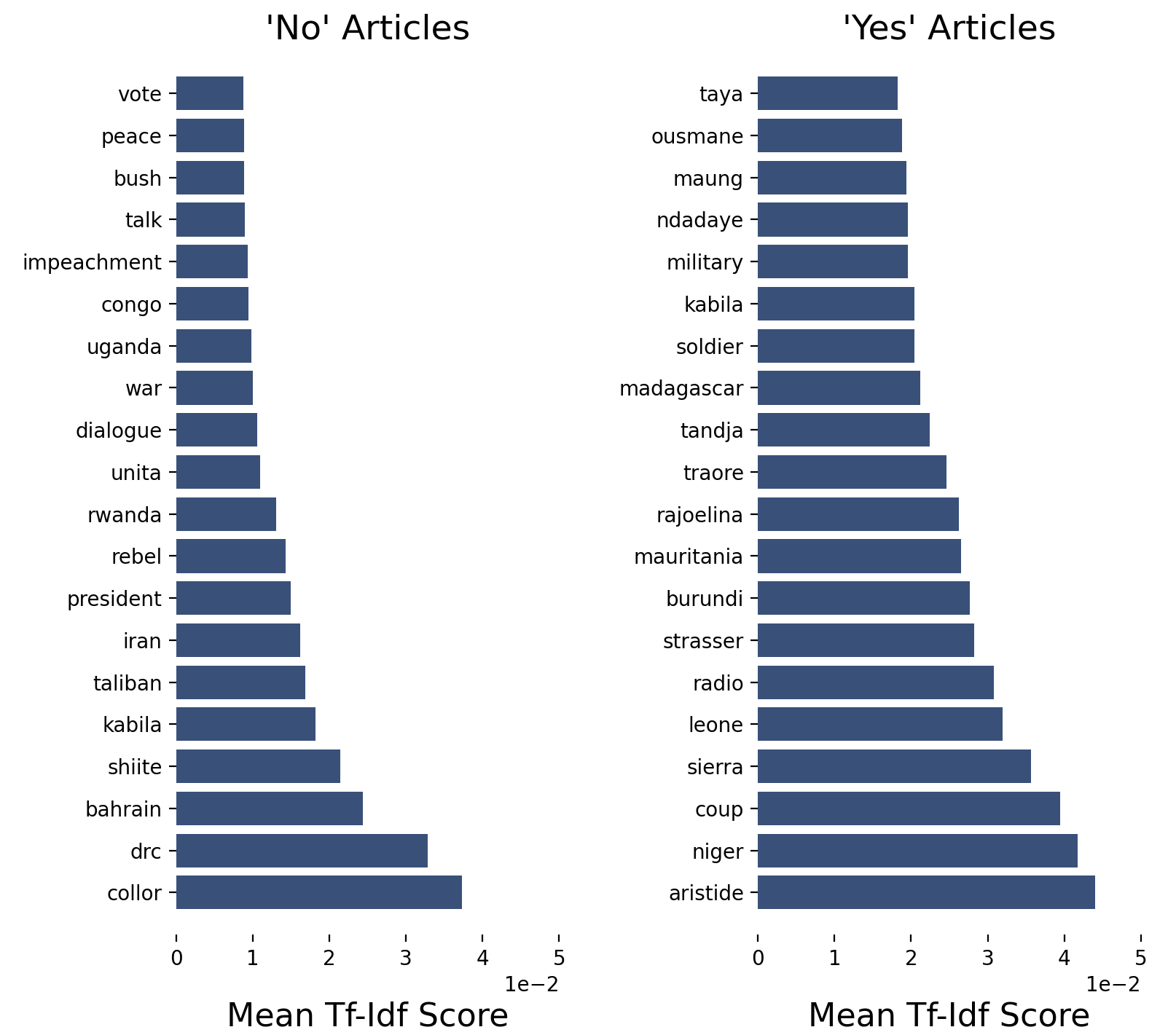}
  \caption{The most significant tokens towards classification of the training set, measured by tf-idf score.}
  \label{figs:TFIDF}
\end{figure}

After conducting exploratory visualizations to gain insights into our corpus and clean it, elements of the ML pipeline can be visualized to better understand how articles are classified. Further preprocessing steps are necessary to prepare the text for feature extraction, which will reduce the article text into its most informative elements for classification.  Here, stop-word removal, parts-of-speech tagging, stemming and lemmatizing procedures are applied to the training set and dependent space corpus.


 \begin{figure}
  \centering
  \includegraphics[width=0.95\linewidth]{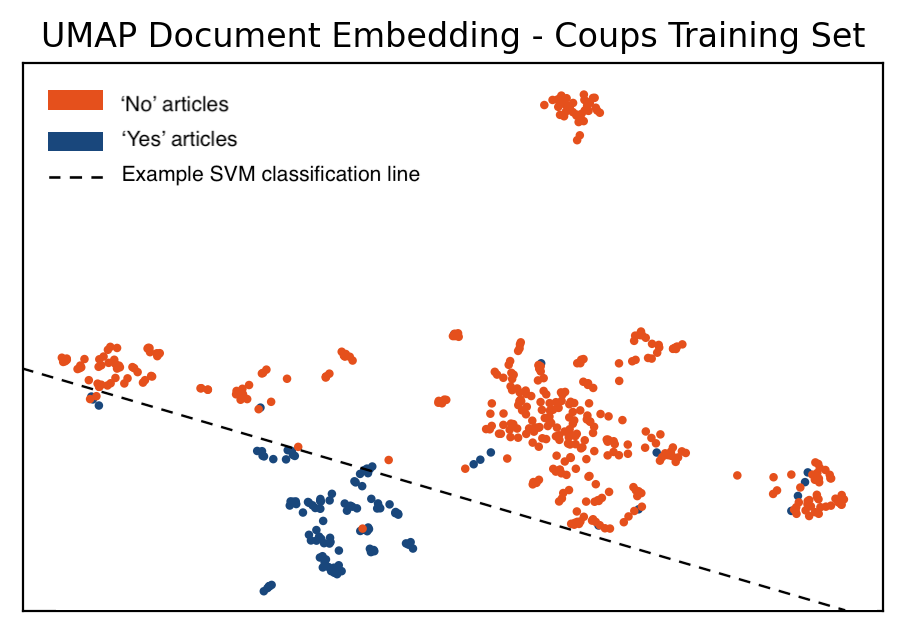}
  \caption{2D projection of the training set documents with an example SVM classification line.}
  \label{figs:UMAP}
\end{figure}

 \begin{figure*}[h]
  \centering
  \includegraphics[width=\textwidth]{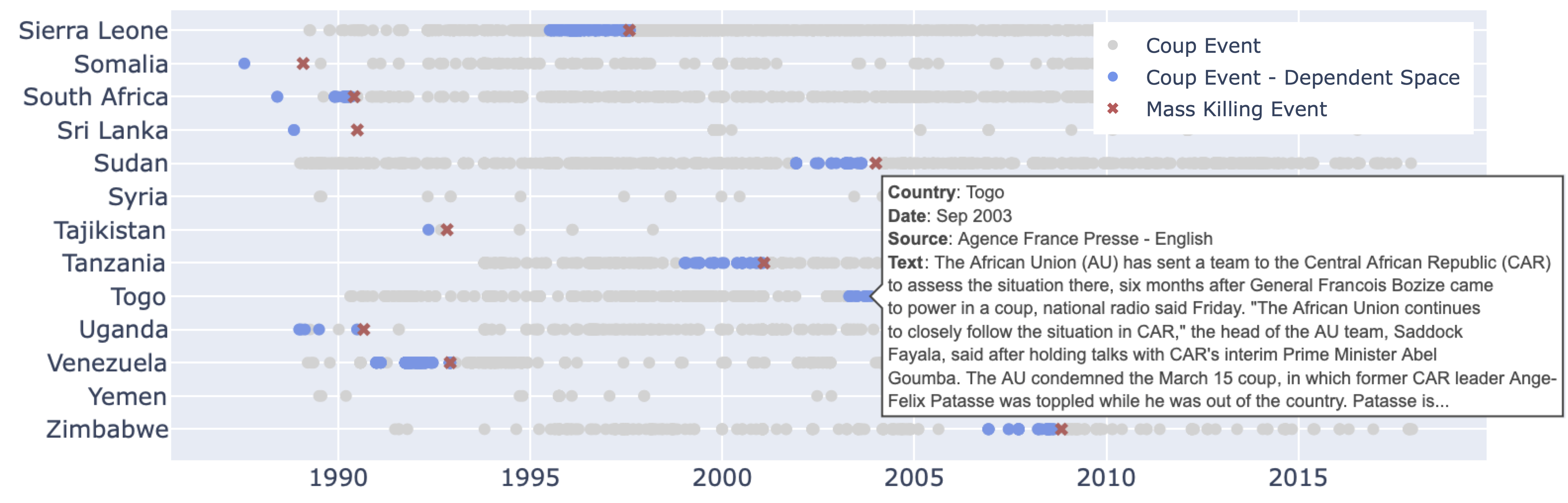}
  \caption{Timeline of extracted articles describing coup events for a subset of countries in the corpus. }
  \label{figs:INFER}
\end{figure*}
 
Visualization of the computed features in the training set offers insight into the linguistic differences between the positive and negative classes.  Our parameter was set to select 5000 features, but this can be adjusted with impacts on classification performance. Figure 3 illustrates the list of top 20 tokens by tf-idf score for each class.  Here, we learn that many proper nouns (country names and leaders of coups in those countries) are weighted as significant indicators that an article is describing a coup event. After classification, this visualization can be repeated for the dependent space corpus and compared to the training set.

The machine learning algorithm will use these feature scores to classify each article, so visualizing the classifier is also useful. Dimensionality reduction techniques such as PCA, t-SNE or UMAP can be used to plot the features into 2-dimensional space and illustrate a classifier. Figure 4 illustrates the use of UMAP \cite{mcinnes2020umap} to generate a projection of the training set.  The true classifier is computed in 5000 dimensions, but an approximate linear classifier demonstrates the class separation.

\section{Classification Validation Visualizations}

After the machine learning tool performs event classification on the unseen articles, visualizing the fully-labeled corpus can help to verify that the classifications make sense in light of any challenges discovered during previous steps. In general, the project hypothesis will determine the type of post-classification visualizations that are most useful in summarizing the classification results.  Because our case study seeks to understand the proximity and dynamics of different triggering events to state-led mass killings, plots like timelines and heatmaps are most suitable to visualize sequences and trends.

For each country, we obtain the list of positively-classified articles from 1989-2017. We distinguish between articles in the two-year window preceding a mass killing event (dependent space corpus in blue) and those at any other time (null space of select countries corpus in grey). Information on coups outside the two-year window allows us to establish controls on their occurrence.

To check the distribution of positively-classified articles for a country across 1989 – 2017, the articles can be visualized on a timeline scatterplot, shown in Figure \ref{figs:INFER}. The plotly library has several features to make visualizations interactive, allowing the user to add details for each article. In cases where many positively-classified articles are plotted in a time period when there were no known coups, this type of interactive visualization is useful to quickly learn about those articles.  

This visualization also revealed a discrepancy in the classifications of the dependent space and select countries corpus: within the two-year window for countries including South Africa, Uganda, and Venezuela, there were instances where positive articles appeared in the select countries corpus but not the dependent space corpus.  This could have happened if the LexisNexis queries did not pull down the same number of articles for the two-year window in each corpus, or if some articles were tagged differently during classification of each corpus. This is possible because the tf-idf vectorizer was fit only on the data set to be classified for each task.  Because tf-idf assigns a weight to each word that is inversely proportional to its frequency across documents in the corpus, the same word can be weighted differently in the dependent space corpus and select countries corpus.  If the differences in weights are large enough, the algorithm may classify the same article differently in each corpus.

\section{Summary}
Visualization techniques are essential to effectively wrangle the complex data and ML tools employed for automated event extraction. The code for generating visualization examples in this paper is publicly available in the ToMK GitHub repository \url{https://github.com/crcresearch/ToMK}.


\bibliographystyle{acl_natbib}
\bibliography{anthology,acl2021,NLPToMK}

\end{document}